\documentclass[final]{cvpr}

\usepackage{times}
\usepackage{epsfig}
\usepackage{graphicx}
\usepackage{amsmath}
\usepackage{amssymb}

\usepackage[pagebackref=false,breaklinks=true,letterpaper=true,colorlinks,urlcolor=blue,citecolor=blue,linkcolor=blue,bookmarks=false]{hyperref}

\def\cvprPaperID{3556} % *** Enter the CVPR Paper ID here
\def\confYear{CVPR 2021}

\def\swseven{0.13\linewidth}

\newcommand{\figdir}{figures}
\begin{document}

%%%%%%%%% TITLE
\title{DeFLOCNet: Deep Image Editing via Flexible Low-level Controls}

\author{Hongyu Liu$^1$ \quad
Ziyu Wan$^{2}$ \quad
Wei Huang$^{3}$ \quad
Yibing Song$^4$\thanks{Y. Song is the corresponding author. The results and code are available at \url{https://github.com/KumapowerLIU/DeFLOCNet}.} \quad
Xintong Han$^1$ \\
Jing Liao$^2$ \quad 
Bing Jiang$^3$ \quad
Wei Liu$^5$ \\
$^{1}$Huya Inc \quad
$^{2}$City University of Hong Kong \quad $^{3}$Hunan University\\
$^{4}$Tencent AI Lab \quad $^{5}$Tencent Data Platform \quad \\
{\tt\small \{liuhongyu1,hanxintong\}@huya.com \quad ziyuwan2-c@my.cityu.edu.hk}\\
{\tt\small yibingsong.cv@gmail.com \quad wl2223@columbia.edu}
}

\maketitle
\pagestyle{empty}  % no page number for the second and the later pages
\thispagestyle{empty} % no page number for the first page
\renewcommand{\tabcolsep}{.5pt}
\begin{figure}[t]
\vspace{-2.75in}
\begin{minipage}{\textwidth}
\begin{center}
\begin{tabular}{ccccccc}
    \vspace{-0.5mm}
    \includegraphics[width=\swseven]{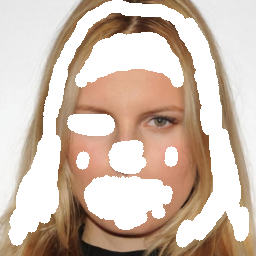}&
    \includegraphics[width=\swseven]{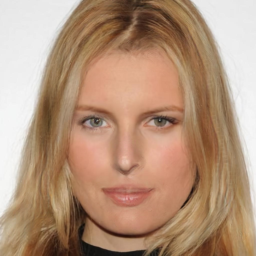}&
    \includegraphics[width=\swseven]{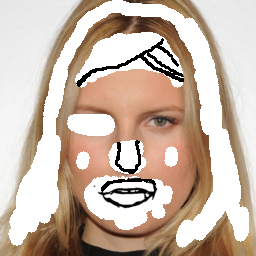}&
    \includegraphics[width=\swseven]{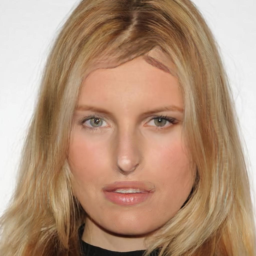}&
    \includegraphics[width=\swseven]{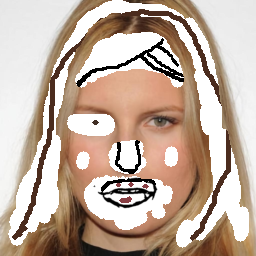}&
    \includegraphics[width=\swseven]{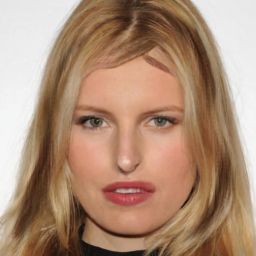}&
    \includegraphics[width=\swseven]{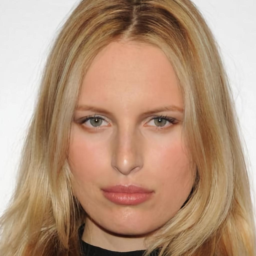}\\
    \vspace{-0.5mm}
    \includegraphics[width=\swseven]{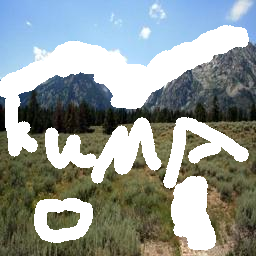}&
    \includegraphics[width=\swseven]{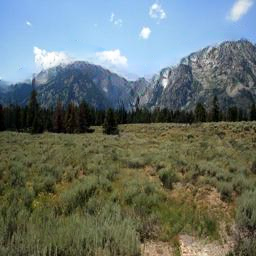}&
    \includegraphics[width=\swseven]{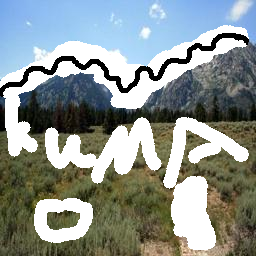}&
    \includegraphics[width=\swseven]{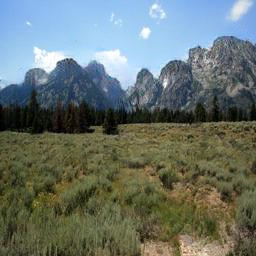}&
    \includegraphics[width=\swseven]{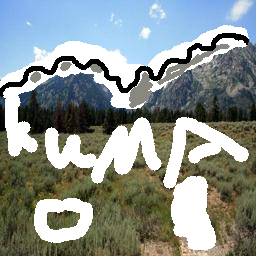}&
    \includegraphics[width=\swseven]{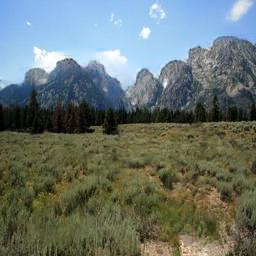}&
    \includegraphics[width=\swseven]{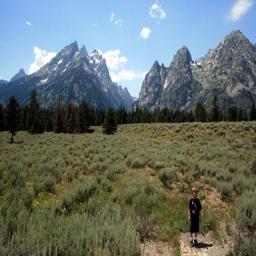}\\
    \label{fig:teaser}
    (a) Input &(b) Ours &(c) Input &(d) Ours &(e) Input &(f) Ours &(g) Original
\end{tabular}
\end{center}
\vspace{-3mm}
\caption{Diversified image editing results under free-form input configurations. There is a face or a natural image with arbitrary hole regions. Without any user inputs shown in (a), our DeFLOCNet automatically fills hole regions shown in (b), which is the same as the image inpainting scenario. Given additional low-level controls (\eg, coarse sketch lines in (c), both lines and colors in (e)), our DeFLOCNet injects these controls directly into structure generation blocks for both user-intended and visually pleasant content creations in (d) and (f).}
\label{fig:teaser}
\end{minipage}
\end{figure}

%%%%%%%%% ABSTRACT
\begin{abstract}
User-intended visual content fills the hole regions of an input image in the image editing scenario. The coarse low-level inputs, which typically consist of sparse sketch lines and color dots, convey user intentions for content creation (\ie, free-form editing). While existing methods combine an input image and these low-level controls for CNN inputs, the corresponding feature representations are not sufficient to convey user intentions, leading to unfaithfully generated content.
In this paper, we propose DeFLOCNet which relies on a deep encoder-decoder CNN to retain the guidance of these controls in the deep feature representations. 
In each skip-connection layer, we design a structure generation block. Instead of attaching low-level controls to an input image, we inject these controls directly into each structure generation block for sketch line refinement and color propagation in the CNN feature space. We then concatenate the modulated features with the original decoder features for structure generation. Meanwhile, DeFLOCNet involves another decoder branch for texture generation and detail enhancement.
Both structures and textures are rendered in the decoder, leading to user-intended editing results.
Experiments on benchmarks demonstrate that DeFLOCNet effectively transforms different user intentions to create visually pleasing content.
\end{abstract}

%%%%%%%%% BODY TEXT

\section{Introduction}
The investigation on image editing is growing as it reduces significant manual efforts during image content generation. Benefiting from the realistic image representations brought by convolutional neural networks (CNNs), image editing is able to create meaningful while visually pleasant content. As shown in Fig.~\ref{fig:teaser}, users can draw arbitrary holes in a natural image as inputs to indicate the regions to be edited. If there are no further inputs given as shown in (a), image editing degenerates to image inpainting, where CNNs automatically fill hole regions by producing coherent image content as shown in (b). If there are additional inputs from users (\eg, lines in (c) and both lines and colors in (e)), CNNs will create meaningful content accordingly while maintaining visual pleasantness. Deep image editing provides flexibility for users to generate diversified content, which can be widely applied in the areas of data enhancement, occlusion removal, and privacy protections.

The flexibility of user controls and the quality of user-intended content generation are challenging to achieve simultaneously in practice. The main difficulty resides on how to transform flexible controls into user-intended content. Existing attempts utilize high-level inputs (\eg, semantic parsing map \cite{hong2018learning}, attributes \cite{perarnau2016invertible}, latent code \cite{Bau:Ganpaint:2019}, language \cite{chen2018language}, and visual context \cite{pathak2016context}) for semantic content generation, but flexibility hinges on the predefined semantics.

On the other hand, utilizing coarse low-level controls (\ie, sketch lines and colors) makes the editing more interactive and flexible. And in this paper, we mainly focus on incorporating such user inputs for image editing, in which we observe two main challenges: (1) Most prior investigations~\cite{yu2019free,jo2019sc,portenier2018faceshop} simply combine an input image and low-level controls together in the image level for CNN inputs. The guidance from these low-level inputs gradually diminishes in the CNN feature space, weakening their influence on generating user-intended contents. Fig.~\ref{fig:qualitative} (c)-(f) show such examples where facial components are not effectively produced, (2) Since users only  provide sparse color strokes to control the generated colors, the model needs to propagate these spatially sparse signals to the desired regions guided by sketches (\ie, colors should fill in the regions indicated by the sketches and not be wrongly rendered across sketch lines) as illustrated in Figs.~\ref{fig:visual} and \ref{fig:blockabl}.

%These methods preliminary attempts by combining an image and low-level controls together in image level for CNN inputs are not satisfying, as shown in Fig.~\ref{fig:qualitative} (b)-(e). This is mainly because the low-level information is significantly dissolved by CNNs during propagation.
%FEAPL \cite{dong2019fashion} adopt the parsing information as reference during editing, and add the low-level controls by normalization layers. However, it can just handle the fashion editing. And the low-level guidance are not well reflected in the final results, since the sketch just inject once at each scale and color cannot be diffused.

\begin{figure*}[t]
\begin{center}
\includegraphics[width=1.0\linewidth]{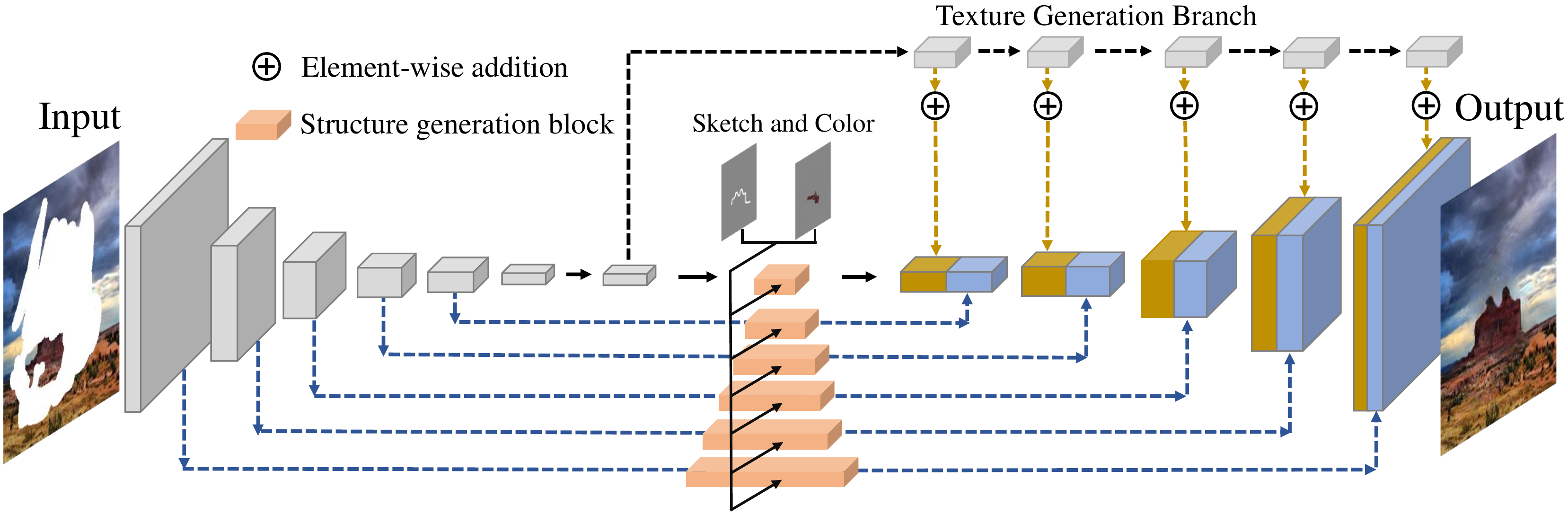}
\caption{An overview of DeFLOCNet. We set each structure generation block within each skip connection layer. The block size corresponds to the level of skip connection layers. Low-level (\ie, free-form) controls are injected into these blocks to modulate encoder features from coarse to fine. The modulated features represent user intentions and supplement decoder features together with texture generation features for output image generation.}
\label{fig:pipeline}
\end{center}
\end{figure*}

To resolve these issues, we propose DeFLOCNet (\ie, Deep image editing via Flexible LO-level Control) to retain the guidance of low-level controls for reinforcing user intentions. Fig. \ref{fig:pipeline} summarizes DeFLOCNet, which is built on a deep encoder-decoder for structure and texture generations on the hole regions. At the core of our contribution is a novel structure generation block (Fig. \ref{fig: editblock} and Sec. \ref{sec:structure}), which is plugged into each skip connection in the network. Low-level controls are directly injected into these blocks for sketch line generation and color propagation in the feature space. The structure features from these blocks are concatenated to the original decoder features accordingly for user-intended structure generation in the hole regions. 

Moreover, we introduce another decoder for texture generation (Sec. \ref{sec:texture}). Each layer of the texture generation decoder is concatenated to the original decoder for texture enhancement. Thus, both structure and texture are effectively produced in the CNN feature space. They supplement original decoder features to bring coarse-to-fine user-intended guidance in the CNN feature space and output visually pleasing editing results.
Experiments on the benchmark datasets demonstrate the effectiveness of our DeFLOCNet compared to state-of-the-art approaches.

%We summarize the contributions of this work as follows:
%\begin{itemize}
%  \item We propose a structure recovery block for image editing. By injecting the low-level inputs into the CNN features, the proposed block can guide the structure reconstruction. 
%  \item We propose a texture decoder to enrich the generated details. With the decoder, the blur and artifacts can be removed effectiveness.
%  \item Extensive experiments on the benchmark datasets show the effectiveness of the proposed editing methods in controlling the generated contents.  The proposed method performs favorably against state-of-the-art editing approaches.
%\end{itemize}

%------------------------------------------------------------------------

\section{Related Work}

{\flushleft \bf Deep Generative Models.}
The advancements in deep generative models~\cite{song2017crest,song2019joint,wang2020rethinking} are inspired by generative adversarial learning~\cite{goodfellow2014generative,song2018vital}. Instead of image generation from random noise, conditioned image generation from inputs activates a series of image translation work. In \cite{isola2017image}, a general framework is proposed to translate semantic labels to natural images. This framework is further improved by using a coarse-to-fine generator and a multi-scale discriminator~\cite{wang2018pix2pixHD}. Besides holistic image generation, subregion image generation (\ie, image inpainting) receives heavy investigations~\cite{yan2018shift, PConv, liu2019coherent}. In contrast to existing image-to-image generation frameworks, our free-form image editing is more flexible to transfer user intentions (\ie, monotonous sketch lines and color dots) into natural image content. 

{\flushleft \bf Image Editing.}
GANs have a lasting influence on image editing development. In \cite{perarnau2016invertible}, Invertible Conditional GANs are proposed to control high-level attributes of generated faces. Then, more effective editing is proposed in \cite{shen2019interpreting} by approximating a disentangled latent space. Semantic parsing maps are utilized in \cite{hong2018learning, gu2019mask,dong2019fashion} as the intermediate representation for guided image editing, while natural language navigates editing in \cite{chen2018language, nam2018text}. Methods based on semantic guidance typically require an explicit correspondence between editing content and semantic guidance. As the semantic guidance is usually fixed with limited options, the editing is thus not flexible (\eg, color and sketch controls). To improve the input flexibility, SC-FEGAN \cite{jo2019sc} proposes to directly combine sketch lines and colors as inputs and send them together with an input image to CNN. Gated convolution is proposed in \cite{yu2019free} for flexibility improvement. As these methods attach user controls directly to input images for CNN input, the influence of user controls diminishes gradually. As a result, limited editing scenarios are supported by these methods (\eg, facial component editing).
Different from existing approaches, we inject low-level controls in the skip connection layers of an encoder-decoder with our structure generation block to gradually reinforce user intentions in a coarse-to-fine manner.

%-------------------------------------------------------------------------
\section{DeFLOCNet}

Fig.~\ref{fig:pipeline} shows an overview of our DeFLOCNet built on top of an encoder-decoder CNN model. The input to the encoder is an image with arbitrary hole regions. Low-level controls, represented by sketch lines and color dots, are sent to the structure generation blocks (SGB) set on the skip-connection layers (Sec. \ref{sec:structure}). Meanwhile, we propose another decoder named texture generation branch (TGB) (Sec. \ref{sec:texture}). The features from SGB and TGB are fused with the original decoder features hierarchically for output image generation.

{\flushleft \bf Motivation}. A vanilla encoder-decoder CNN may be prevalent for image generation while it is insufficient to recover missing content in the hole regions which are almost empty with sparse low-level controls. This is due to the limited ability of the encoder features to represent both natural image contents and free-form controls. To improve feature representations, we only send images to the encoder while injecting controls multiple times into all the skip connection layers via SGB. Consequently, user intentions are reinforced continuously via feature modulations. The reinforced features, together with the texture generation features, supplement the original decoder features in a coarse-to-fine fashion to generate both user-intended and visually pleasant visual content.

\begin{figure*}[t]
\begin{center}
\begin{tabular}{c}
\includegraphics[width=0.85\linewidth]{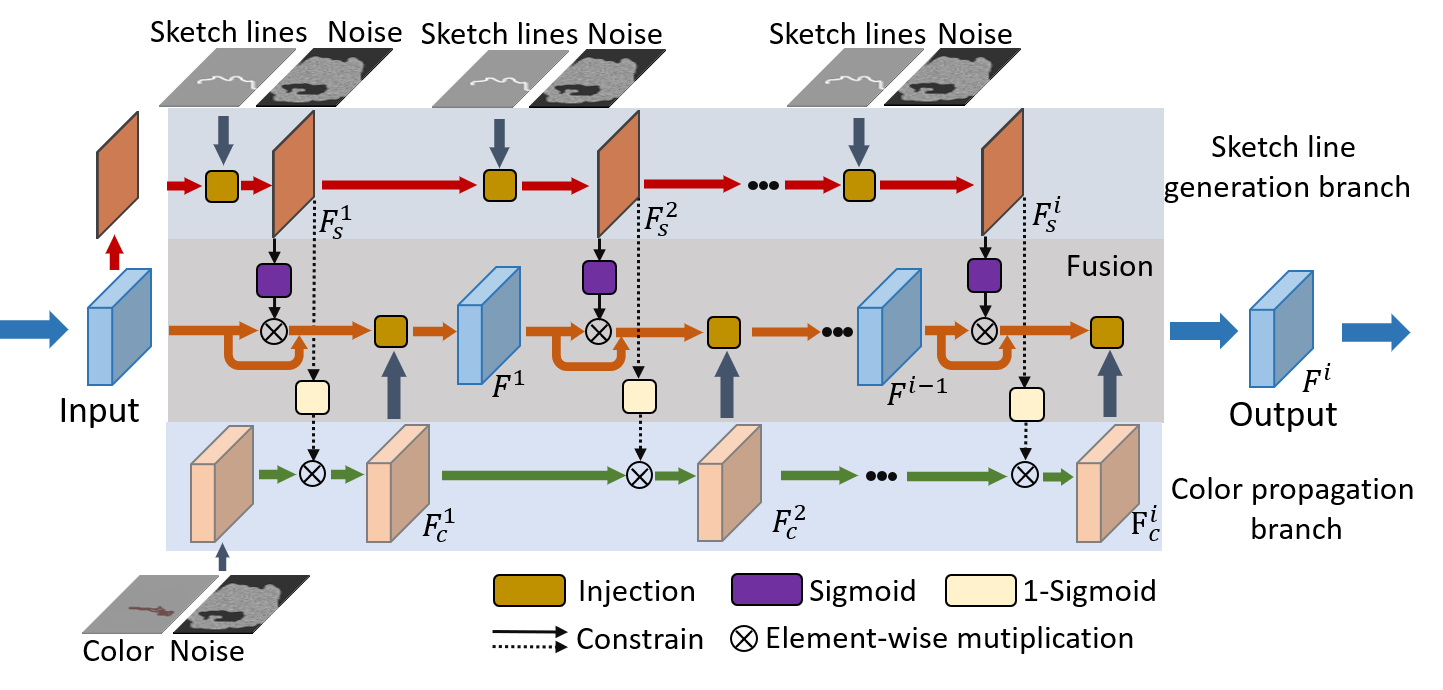}\\
\end{tabular}
\caption{The architecture of a structure generation block where $i\in[1,2,...,6]$ denotes its size. The size of an SGB increases (\ie, $i$ increases) when it is integrated into a shallower encoder layer. An SGB consists of one sketch line generation branch, one color propagation branch, and a fusion branch. The sketch lines are gradually refined to guide the color propagation. These features are fused with the input image features for content generation in the hole region. }
\label{fig: editblock}
\end{center}
\end{figure*}

%-------------------------------------------------------------------------
\subsection{Structure Generation}
\label{sec:structure}

We inject low-level controls into SGBs to modulate features of user intention reinforcement. Fig.~\ref{fig: editblock} shows the architecture of one SGB, which consists of three
branches for progressive sketch line generation, color propagation, and feature fusion, respectively. The size of one SGB increases when it is integrated into a shallower encoder layer, since the shallower is close to the image level and we need stronger low level to guide the feature generation. The sketch line generation branch repeatedly injects and refines the sketch generation, avoiding the user control diminishing phenomenon. Then, the features of sketch are utilized in the color propagation branch to regularize the color to fill in the desired regions. Finally, the fusion branch injects the sketch and color features into the original feature to produce output editing results. 

{\flushleft \bf Control injection.}
We first introduce a control injection operation, which is a basic building block in our SGB. The control operation follows \cite{spade}. Specifically, we denote $F^{\rm in} \in \mathbb{R} ^{H\times W\times C}$ as the input feature map and $L$ as the information we want to inject into $F^{\rm in}$. %The sizes of $F^{\rm in}$ and $L$ are $H\times W\times C$ and $H\times W\times 3 $, respectively.
Suppose the injection operation is $\mathcal{I}(\cdot)$, and the element in the injected feature $F^{\rm out}_{x,y,c}=\mathcal{I}(F^{\rm in}_{x,y,c}, L) $ can be obtained by:
\begin{equation}\label{eq:inj}
   \mathcal{I}(F^{\rm in}_{x,y,c}, L) =\gamma_{x, y, c}(L)\cdot \frac{F^{\rm in}_{x,y,c}-\mu_{c}}{\sqrt{\sigma_{c}^{2}+\epsilon}}+ \beta_{x, y, c}(L),
\end{equation}
where $\gamma_{x, y, c}(L)$ and $\beta_{x, y, c}(L)$ are two variables controlling the influence from $L$ in element-wise precision, $\mu_{c} = \frac{1}{H\times W} \sum_{x=1}^{H} \sum_{y=1}^{W} F^{\rm in}_{ x, y, c}$, and  $\sigma_{c}=\sqrt{\frac{1}{H\times W} \sum_{x=1}^{H} \sum_{y=1}^{W}\left(F^{\rm in}_{x,y,c}-\mu_{c}\right)^{2}}$. In this paper, we use two convolutional layers to generate $\gamma_{x, y, c}(L)$ and $\beta_{x, y, c}(L)$ at each element location. As a result, low-level controls are mapped into the feature space to correlate to the input feature maps.

{\flushleft \bf Sketch line generation}.
Given an input feature $F^{in}$, the sketch line generation branch first performs an element-wise average along the channel dimension to produce a single-channel feature map $F^{in}_{avg}$. We denote the sketch image as $S$, a random noise image as $N$, and a mask image containing a hole region as $M$. The output feature after control injection can be written as:
\begin{equation}
    F_s=\mathcal{I}(F^{\rm in}_{avg},S\oplus(N\odot M)),
\end{equation}
where $F_s$ is the injected feature, $\oplus$ is the concatenation operator, and $\odot$ is the element-wise multiplication operator. In practice, a single injection does not generate recovered sketch lines completely. We use several injections in the sketch line generation branch to progressively refine sketch lines. The injection in the $i$-th skip connection is:
\begin{equation}
    F_s^{i}=\mathcal{I}(\operatorname{Conv}(F_s^{i-1}),S\oplus(N\odot M)),
\end{equation}
where the output features from the previous injection $F_s^{i-1}$ are passed through a convolutional layer for the current injection input. During training, we use the ground truth sketch lines extracted from the original images. When editing images, we adopt user inputs for sketch line refinement.

\begin{figure}[t]
    \begin{center}
    \includegraphics[width=0.8\linewidth]{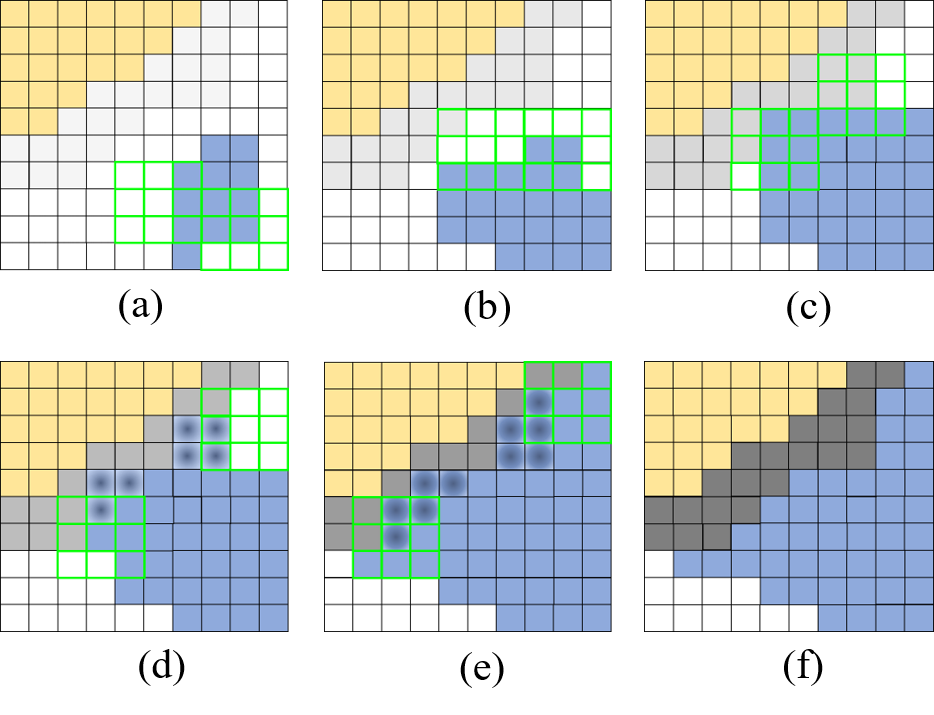}
    \end{center}
    \vspace{-2mm}
    \caption{Sketch line refinement guides color propagation. The green window indicates the regular convolution operation, which is denoted by $\text{Conv}$ in Eq.~\ref{eq:color}. The blue color diffuses along all directions in (a)-(b) when sketch lines (gray lines) are initially refined. Color propagation is weakened via $1-\sigma$ around the contour lines that are gradually completed as shown in (c)-(e). The blue dots indicate the weakened elements along the lines. The propagation result is shown in (f) where blue color propagation follows contour lines.}
    \label{figure: diffusion}
\end{figure}

{\flushleft \bf Color propagation.}
We propose a color propagation branch in parallel to the sketch generation branch. In order to guide the color propagation via sketch lines, we use injected features $F_s^i$ from the sketch line generation branch. The guiding process can be written as:
\begin{equation}
    F_{c}^{i}=(1-\sigma (F_s^{i-1}))\otimes \operatorname{Conv}(F_{c}^{i-1}),
\label{eq:color}
\end{equation}
where $F_c^i$ is the color features guided by $F_s^{(i-1)}$ and $\sigma$ is the sigmoid activation function.

Fig.~\ref{figure: diffusion} illustrates how color propagates under sketch guidance. In (a) and (b), the sketch lines in gray are not recovered well and the blue color tends to diffuse in all directions. As the sketch lines are gradually refined to complete contours as shown in (c)-(e), the blue color does not penetrate the contour lines via the consecutive $\sigma$ and $1-\sigma$ operations. Finally, blue color propagates along the contour lines without penetration as shown in (f).

{\flushleft \bf Fusion.}
The features from the sketch and color branches are fused together via the injection operation as follows:
\begin{equation}
    F^{i}=\mathcal{I}(F^{i-1}\otimes(\sigma (F_s^{i-1})+\mathbf{1}),F_{c}^{i-1}),
\label{eq:f}
\end{equation}
where $F^i$ is the fused feature. We set different numbers of injection operations in the fusion branch for each scale, respectively. For the skip-connection from the initial encoder layer where features are in large resolution, we employ 6 injection operations in the fusion branch. We gradually decrease this number to 1 on the skip connection layers from the last encoder layer.

%-------------------------------------------------------------------------
\subsection{Texture Generation}
\label{sec:texture}
We use SGB to reinforce user intentions during hole filling in the CNN feature space. The modulated features represent structure content while texture representation is limited. This is partially because low-level controls injected into the skip-connection layers do not contain sufficient texture guidance. Meanwhile, the sketch lines attend the encoder features to structure content rather than textures.% The texture generation shall be independent to the structure generation during the image editing process.

As encoder features do not relate to low-level controls, we propose a texture generation branch that takes the features from the last encoder layer as input. Fig.~\ref{fig:pipeline} shows how TGB is integrated into the current pipeline. The architecture of TGB is the same as that of the original decoder. We add the feature maps from each layer of TGB to the corresponding decoder features for texture generation. TGB supplements  decoder  features  via  residual  aggregations. As  structure  features  are learned via SGB, TGB will focus on the features representing region details.  The enriched decoder features are then concatenated with the features from SGBs for output generation where there are both structures and textures in the hole regions.

%-------------------------------------------------------------------------
\subsection{Objective Function}

We utilize several objective loss functions to train DeFLOCNet in an end-to-end fashion. These functions include pixel reconstruction loss, perceptual loss~\cite{johnson-eccv16-perceptual}, style loss~\cite{styleloss}, relativistic average LS adversarial loss~\cite{Alexia-ICLR19-rslgan12}, and total variation loss~\cite{PConv}. During training, we extract the sketch lines and color in the hole regions. We denote $I_{\rm out}$ as the output result and $I_{\rm gt}$ as the ground truth ($I_{\rm out}$ and $I_{\rm gt} \in\mathbb{R}^{H \times W \times 3}$). The loss terms can be written as follows:

{\flushleft \bf Pixel reconstruction loss.} We measure the pixel-wise difference between $I_{\rm gt}$ and $I_{\rm gt}$ as:
\begin{equation}\label{eq:O2 Reconstruct Loss}
    L_{\rm re} = \lVert I_{\rm out}-I_{\rm gt} \rVert_1 ,
\end{equation}

{\flushleft \bf Perceptual loss.} We consider high-level feature representation and human perception to utilize the perceptual loss, which is based on the ImageNet-pretrained VGG-16 backbone. The perceptual loss can be written as:
\begin{equation}\label{eq:prec}
\begin{aligned}
L_{\rm prec}= \sum_i \frac{1}{N_i} \lVert F_{i}(I_{\rm out})-F_{i}(I_{\rm gt}) \rVert_1 ,
\end{aligned}
\end{equation}
where $F_{i}$ is the feature map of the $i$-th layer of the VGG-16 backbone and, $N_i$ is the number of elements in $F_{i}$. In our work, $F_{i}$ corresponds to the activation maps from layers ReLu1\_1, ReLu2\_1, ReLu3\_1, ReLu4\_1, and ReLu5\_1.

{\flushleft \bf Style loss.} The transposed convolutional layers of the decoder will bring checkerboard effect~\cite{PConv}, which can be mitigated by the style loss. Suppose the size of feature map $F_i$ is $C_i\times H_i\times W_i$. We write the style loss as:
\begin{equation}\label{eq:style}
\begin{aligned}
L_{\rm style}=   \sum_i \frac{1}{M_i} \lVert G_i^F (I_{\rm out})- G_i^F (I_{\rm gt})\rVert_1 ,
\end{aligned}
\end{equation}
where $G_i^F$ is a $C_i \times C_i$ Gram matrix computed based on the feature maps, and $M_i$ is the number of elements in $G_i^F$. These feature maps are the same as those used in the perceptual loss as illustrated above.

{\flushleft \bf Relativistic average LS adversarial loss.}
We utilize global and local discriminators for perception enhancement. The relativistic average LS adversarial loss is adopted for our discriminators, which can be written as:
\begin{equation}\label{eq:LR}
\begin{aligned}
\begin{gathered}
L_{\rm adv}=-\mathbb{E}_{x_{r}}[\operatorname{log}(1-D_{ra}(x_r,x_f))]-\\
\mathbb{E}_{x_{f}}[\operatorname{log}(D_{ra}(x_f,x_r))],
\end{gathered}
\end{aligned}
\end{equation}
where $D_{ra}(x_r,x_f)=\operatorname{sigmoid}(C(x_r)-\mathbb{E}_{x_{f}}[C(x_f)])$ and $C(\cdot)$ indicates the local or global discriminator, and real and fake data pairs $(x_r,x_f)$ are sampled from $I_{\rm gt}$ and $I_{\rm out}$.

{\flushleft \bf Total variation loss.} This loss is set to
add the smoothness penalty on the generated regions, which is defined as:
\begin{equation}\label{eq:TV}
\begin{gathered}
\begin{aligned}
L_{\rm tv}= \sum_{(i,j)\in R,(i,j+1)\in R } \frac{\lVert I_{\rm out}^{i,j+1}-I_{\rm out}^{i,j} \rVert_1}{N} + \\
\sum_{(i,j)\in R,(i+1,j)\in R } \frac{ \lVert I_{\rm out}^{i+1,j}-I_{\rm out}^{i,j} \rVert_1}{N}, 
\end{aligned}
\end{gathered}
\end{equation}
where $N$ is the number of elements in $I_{\rm out}$, and $R$ denotes the hole regions.

{\flushleft \bf Total losses.}
The whole objective function of DeFLOCNet can be written as:
\begin{equation}\label{eq:final loss}
\begin{aligned}
L_{\rm Total}=&\lambda_1\cdot L_{\rm re}+\lambda_2\cdot L_{\rm prec}+\lambda_3\cdot L_{\rm style}+\\&\lambda_4\cdot L_{\rm tv}+\lambda_5\cdot L_{\rm adv},
\end{aligned}
\end{equation}
where $\lambda_1$, $\lambda_2$, $\lambda_3$, $\lambda_4$ and $\lambda_5$ are the scalars controlling the influence of each loss term. We empirically set $\lambda_1=1$, $\lambda_2=0.05$, $\lambda_3=250$, $\lambda_4=0.1$ and $\lambda_5=0.1$.

\def\sweight{0.12\linewidth}
\begin{figure*}[t]
\begin{center}
\begin{tabular}{cccccccc}
\includegraphics[width=\sweight]{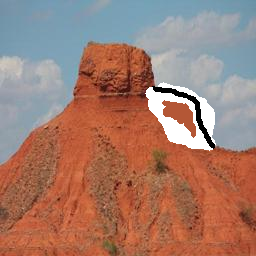}&
\includegraphics[width=\sweight]{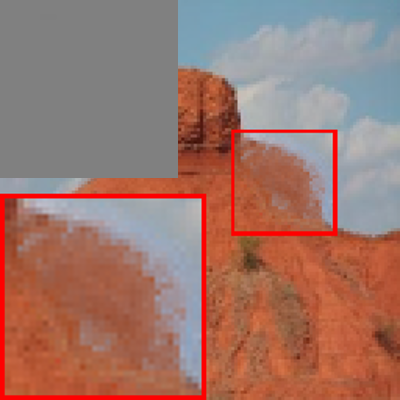}&
\includegraphics[width=\sweight]{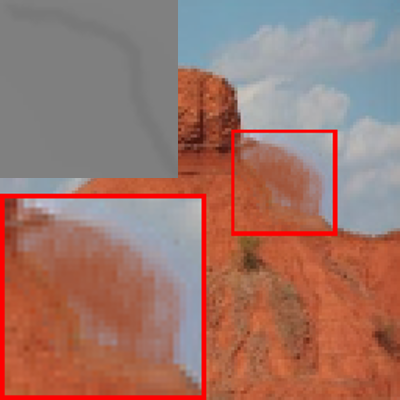}&
\includegraphics[width=\sweight]{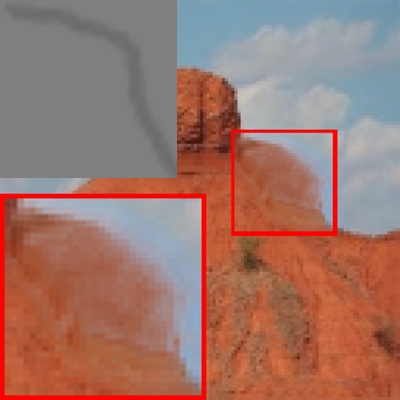}&
\includegraphics[width=\sweight]{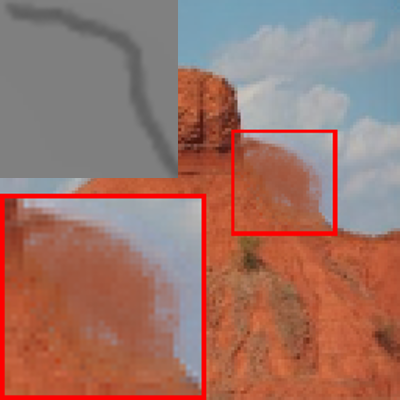}&
\includegraphics[width=\sweight]{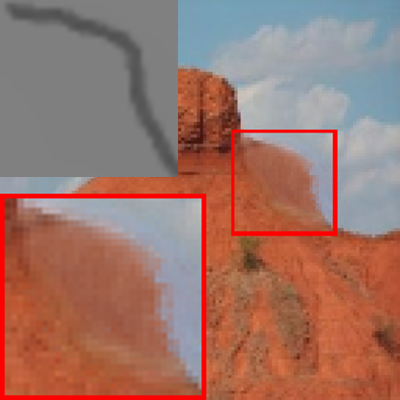}&
\includegraphics[width=\sweight]{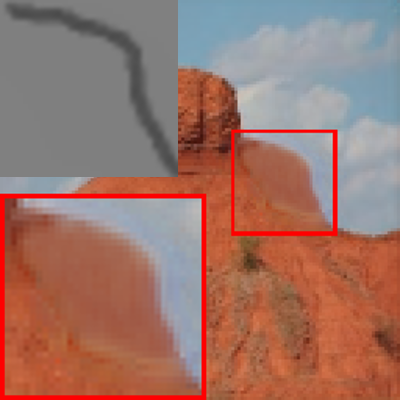}&
\includegraphics[width=\sweight]{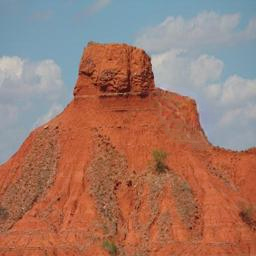}\\
(a) Input& (b) $F^1$& (c) $F^2$& (d) $F^3$& (e) $F^4$& (f) $F^5$& (g) $F^6$& (h) Output\\
\end{tabular}
\end{center}
\vspace{-2mm}
\caption{Feature map visualizations. We map $F^i$ to color images via a $1\times1$ convolutional layer and show $F_s^i$ accordingly at the top left corner. The color propagates according to the sketch line refinement, which validates the illustration in Fig.~\ref{figure: diffusion}. }
\label{fig:visual}
\end{figure*}

\subsection{Visualizations}
The feature representations of input sketch lines are gradually enriched to become those of contours in one SGB. The enriched lines guide the color propagation process for edge-preserving feature modulation. To validate this effect, we visualize the feature maps from the fusion branch of one SGB set in the coarse level. Following the visualization techniques in \cite{Liu2019MEDFE}, we visualize the 6 injection operations in the fusion branch. Specifically, we use a $1\times1$ convolutional layer to map each CNN feature $F^i$ ($i\in[1,2,...,6]$) to one color image, and another $1\times1$ convolutional layer to map each CNN feature $F^i_s$ to one grayscale image. The  weights  of  the mapping convolutions  are  learned. The content shown in the  images indicates the corresponding feature representations in SGB.

Fig.~\ref{fig:visual} shows the visualization results. The input image is shown in (a) where the low-level controls are sent to SGBs. The visualization of $F^i$ is shown in (b)-(g) where the features $F^i_s$ from the sketch generation branch are shown at the top left corner, respectively. We observe that during initial sketch line generation shown in (b)-(c), color propagates in all directions. When sketch lines are gradually completed as shown in (d)-(g), color propagates along these lines to formulate a clear boundary. These feature maps $F^i$ are then concatenated to the original decoder for image structure generation shown in (h). The visualization of color propagation is similar to that in Fig.~\ref{figure: diffusion}, where the $1-\sigma$ operation is effective to prevent the color diffusions.

\section{Experiments}
We evaluate on a natural image dataset Places2~\cite{zhou2017places} and a face image dataset CelebA-HQ~\cite{CelebAMask-HQ}. During training, we follow PConv~\cite{PConv} and HED~\cite{HED} to create hole regions and input sketch lines, respectively. The color inputs for face images are from GFC \cite{GFC} and for natural images are from RTV \cite{RTV}. During training, we choose arbitrary hole regions and low-level controls. Adam optimizer~\cite{kingma2014adam} is adopted with a learning rate of $2\times10^{-4}$. The training epoches for CelebA-HQ and Place2 datasets are 120 and 40, respectively. The resolution of  synthesized  images is 256$\times$256. All the experiments are conducted on one Nvidia 2080 Ti GPU. The various edges of training images ensure our method to handle diverse and deformed strokes.
%The distribution consistency between the edges used for training and testing is guaranteed as we randomly sample all the image regions to utilize different forms of edges during training.

\begin{figure*}[t]
    \begin{center}
    \small
\begin{tabular}{ccccccc}
\vspace{-0.5mm}\includegraphics[width=\swseven]{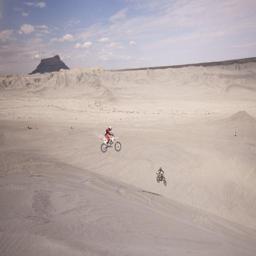}&
\includegraphics[width=\swseven]{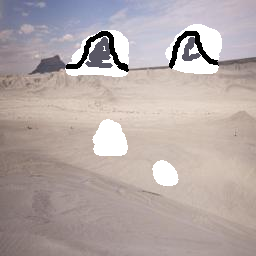}&
\includegraphics[width=\swseven]{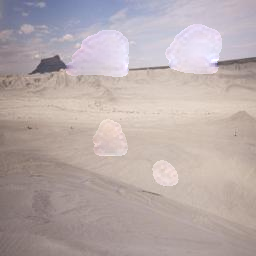}&
\includegraphics[width=\swseven]{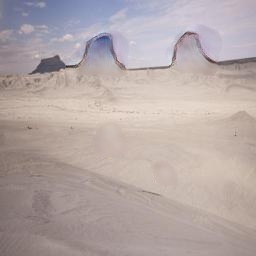}&
\includegraphics[width=\swseven]{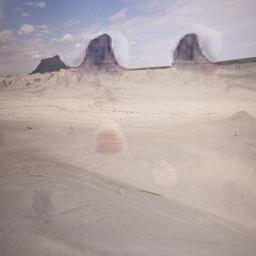}&
\includegraphics[width=\swseven]{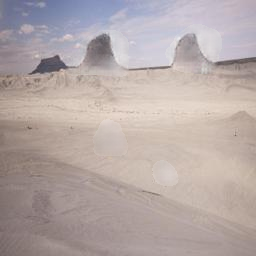}&
\includegraphics[width=\swseven]{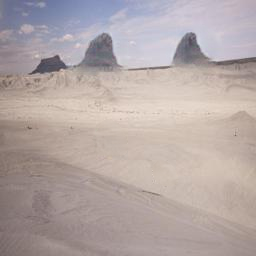}\\
\vspace{-0.5mm}\includegraphics[width=\swseven]{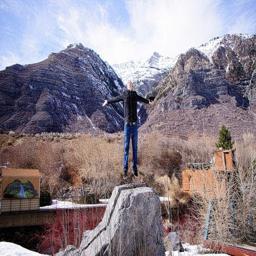}&
\includegraphics[width=\swseven]{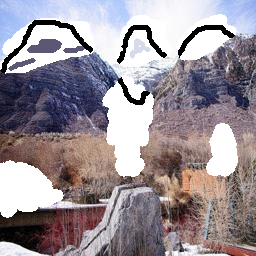}&
\includegraphics[width=\swseven]{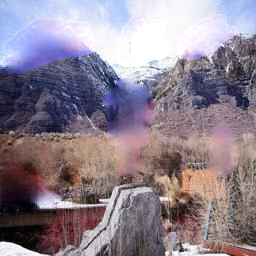}&
\includegraphics[width=\swseven]{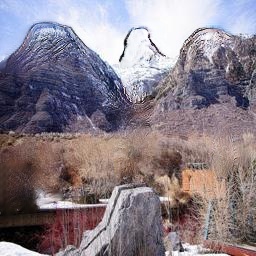}&
\includegraphics[width=\swseven]{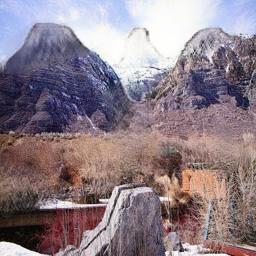}&
\includegraphics[width=\swseven]{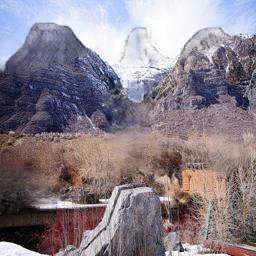}&
\includegraphics[width=\swseven]{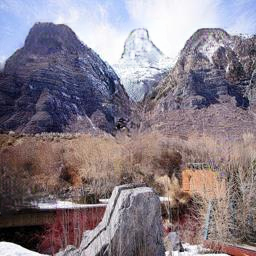}\\
\vspace{-0.5mm}\includegraphics[width=\swseven]{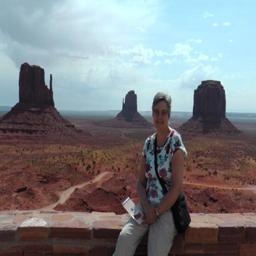}&
\includegraphics[width=\swseven]{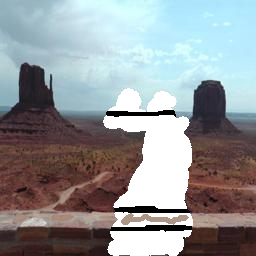}&
\includegraphics[width=\swseven]{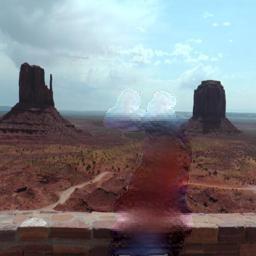}&
\includegraphics[width=\swseven]{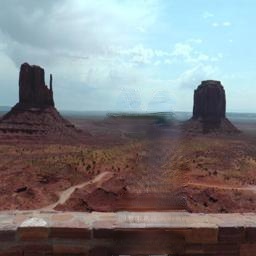}&
\includegraphics[width=\swseven]{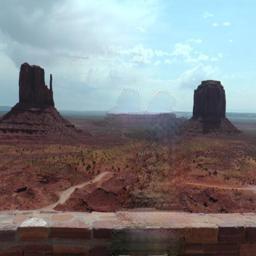}&
\includegraphics[width=\swseven]{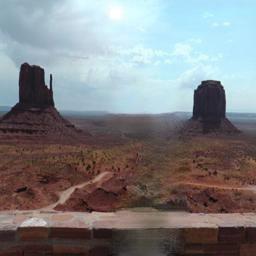}&
\includegraphics[width=\swseven]{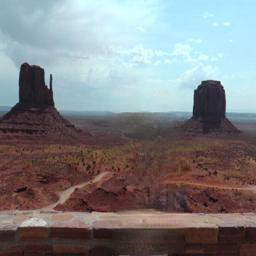}\\
\vspace{-0.5mm}\includegraphics[width=\swseven]{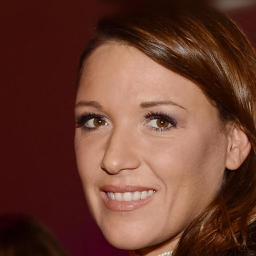}&
\includegraphics[width=\swseven]{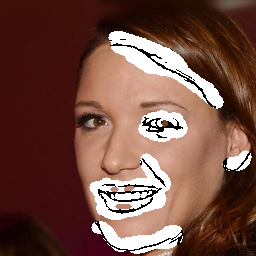}&
\includegraphics[width=\swseven]{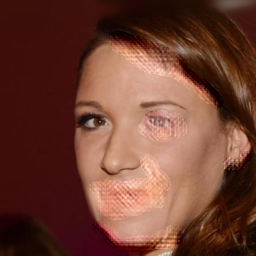}&
\includegraphics[width=\swseven]{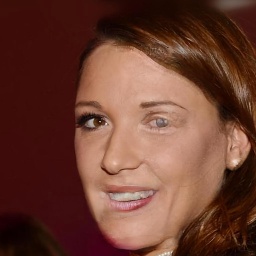}&
\includegraphics[width=\swseven]{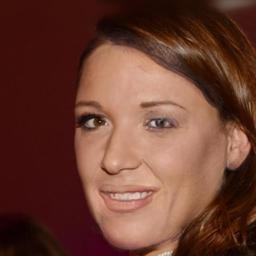}&
\includegraphics[width=\swseven]{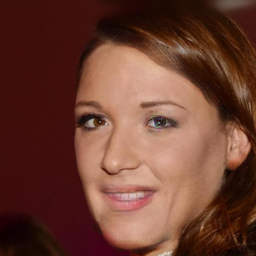}&
\includegraphics[width=\swseven]{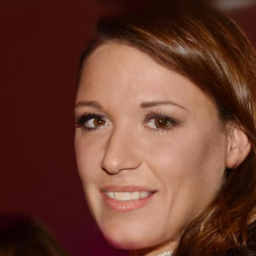}\\
\vspace{-0.5mm}\includegraphics[width=\swseven]{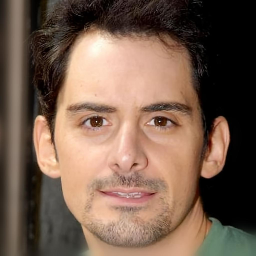}&
\includegraphics[width=\swseven]{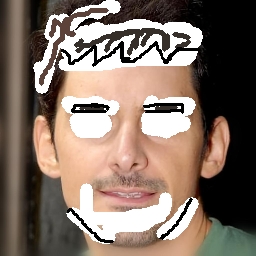}&
\includegraphics[width=\swseven]{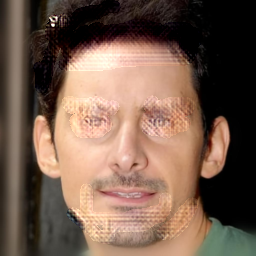}&
\includegraphics[width=\swseven]{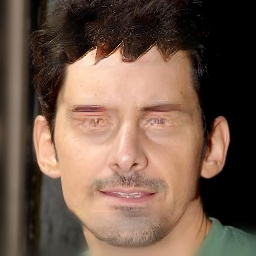}&
\includegraphics[width=\swseven]{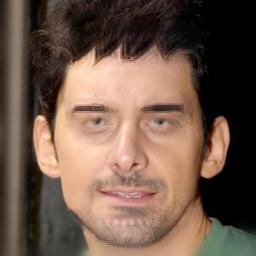}&
\includegraphics[width=\swseven]{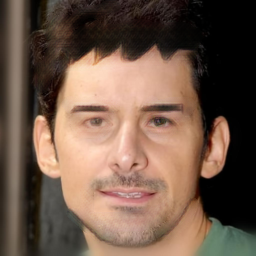}&
\includegraphics[width=\swseven]{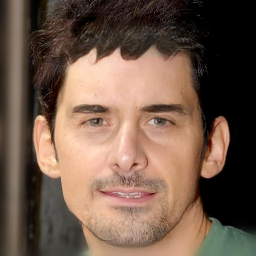}\\
\vspace{-0.5mm}\includegraphics[width=\swseven]{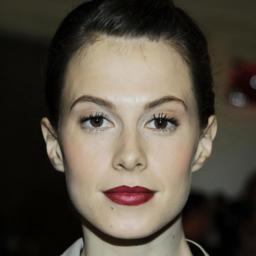}&
\includegraphics[width=\swseven]{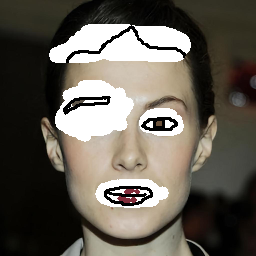}&
\includegraphics[width=\swseven]{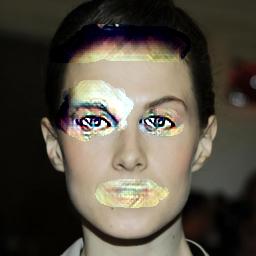}&
\includegraphics[width=\swseven]{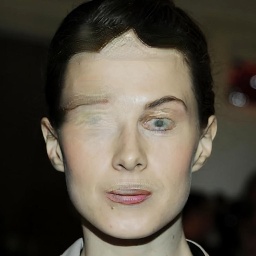}&
\includegraphics[width=\swseven]{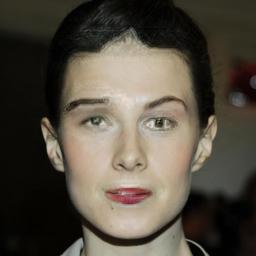}&
\includegraphics[width=\swseven]{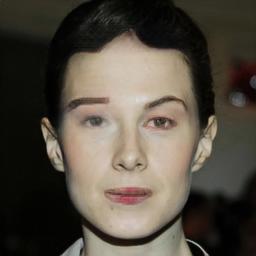}&
\includegraphics[width=\swseven]{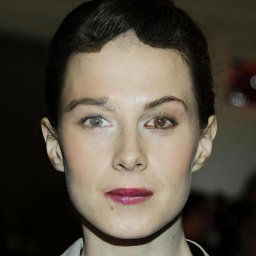}\\
(a) Original & (b) Input &(c) PConv~\cite{PConv} &(d) DF2~\cite{yu2019free}  & (e) P2P~\cite{isola2017image} & (f) FEGAN~\cite{jo2019sc}  &(g) Ours \\
\end{tabular}
\end{center}
\vspace{-2mm}
\caption{Visual comparisons with state-of-the-art methods. Original images are in (a). Input images are in (b) with low-level controls in the hole regions. Our method is  effective to generate user-intended and visually pleasant contents in (g).}
\label{fig:qualitative}
%\vspace{-2em}
\end{figure*}

\subsection{State-of-the-art Comparisons}

We evaluate existing editing methods including SC-FEGAN \cite{jo2019sc}, Partial Conv \cite{PConv}, Deepfill2 \cite{yu2019free} and Pix2Pix \cite{isola2017image}. All these methods are retrained using the official implementations on the same datasets with the same input configurations for fair comparisons. The only difference between DeFLOCNet and other methods is that we send low-level controls into skip-connection layers rather than the encoder. 

{\flushleft \bf Visual Evaluations.}
Fig.~\ref{fig:qualitative} shows the visual comparison results. The original clean images are shown in (a). The input of existing methods is shown in (b). For a straight-forward display, we combine input images with masks and low-level controls together. The results produced by Partial Conv and Deepfill2 are shown in (c) and (d) where structure distortions and blurry textures exist. This is because these two methods tend to incorporate neighboring content when filling the hole regions. They are not effective to generate meaningful content given by the users. In comparison, the results generated by Pix2Pix and SC-FEGAN are improved as shown in (e) and (f). These two methods focus more on the user controls and utilize adversarial learning to generate perceptual realistic contents. However, as these methods attach user controls to color images for the network input, structure generation is thus limited. 
The feature representations are not sufficient to convey both color images and low-level controls since their data distributions are extremely unbalanced. The structures of the eye regions shown in the last three rows are not effectively generated in (e) and (f).

Unlike existing methods that combine all the inputs together, DeFLOCNet sends the color image into the encoder and low-level controls into skip-connection layers via SGBs. These controls gradually enrich encoder features in each skip-connection layer, and refine these features from coarse to fine across multiple skip-connection layers. The results of our method are shown in (g) where image content is effectively generated with detailed textures.

\renewcommand{\tabcolsep}{5pt}
\begin{table*}[t]
\begin{center}
\caption{Numerical evaluations on CelebA-HQ and Places2 datasets. $\downarrow$ indicates lower is better while $\uparrow$ indicates higher is better.}
\begin{tabular}{l|c|c|c|c|cccccc}
\hline
      & \multicolumn{5}{c|}{CelebA-HQ}                                                                                    & \multicolumn{5}{c}{Places2}                                                                                                \\ \hline
Model & \begin{tabular}[c]{@{}c@{}}PConv\\ \cite{PConv}\end{tabular}   & \begin{tabular}[c]{@{}c@{}}DF2\\ \cite{yu2019free}\end{tabular}   & \begin{tabular}[c]{@{}c@{}}SC-\\ FEGAN \cite{jo2019sc}\end{tabular} & \begin{tabular}[c]{@{}c@{}}P2P\\ \cite{isola2017image}\end{tabular}  & \multicolumn{1}{l|}{Ours}           & \multicolumn{1}{c|}{\begin{tabular}[c]{@{}c@{}}PConv\\ \cite{PConv}\end{tabular}}  & \multicolumn{1}{c|}{\begin{tabular}[c]{@{}c@{}}DF2\\ \cite{yu2019free}\end{tabular}}    & \multicolumn{1}{c|}{\begin{tabular}[c]{@{}c@{}}SC-\\ FEGAN \cite{jo2019sc}\end{tabular}} & \multicolumn{1}{c|}{\begin{tabular}[c]{@{}c@{}}P2P\\ \cite{isola2017image}\end{tabular}}   & Ours           \\ \hline
PSNR$\uparrow$  & 22.12 & 24.48 & 24.56                                               & 25.17 & \multicolumn{1}{l|}{\textbf{25.42}} & \multicolumn{1}{c|}{20.14}  & \multicolumn{1}{c|}{22.0}   & \multicolumn{1}{c|}{23.57}                                               & \multicolumn{1}{c|}{21.90} & \textbf{24.30} \\ \hline
SSIM$\uparrow$  & 0.84  & 0.89  & 0.88                                                & 0.89  & \multicolumn{1}{c|}{\textbf{0.90}}  & \multicolumn{1}{c|}{0.62}   & \multicolumn{1}{c|}{0.67}   & \multicolumn{1}{c|}{0.75}                                                & \multicolumn{1}{c|}{0.68}  & \textbf{0.77}  \\ \hline
FID$\downarrow$   & 24.34 & 23.88 & 14.63                                               & 12.45 & \multicolumn{1}{c|}{\textbf{9.92}}  & \multicolumn{1}{c|}{201.09} & \multicolumn{1}{c|}{113.54} & \multicolumn{1}{c|}{77.82}                                               & \multicolumn{1}{c|}{93.46} & \textbf{63.56} \\ 
\hline
Human$\uparrow$   & 0.2\%    & 0.8\%   & 21.3\%  & 15.1\%                            & \multicolumn{1}{c|}{\textbf{54.2\%}}  & \multicolumn{1}{c|}{0.0\%}   & \multicolumn{1}{c|}{3.7\%}   & \multicolumn{1}{c|}{18.2\%}   & \multicolumn{1}{c|}{4.5\%}       & \textbf{73.6\%} \\ \hline
\end{tabular}
\label{tab:all}
\end{center}
\end{table*}

% \renewcommand{\tabcolsep}{2pt}
% \begin{table}
% \centering
%  \caption{Human Subject Evaluation. Each subject selects the most visually pleasant result.}
%  \vspace{2mm}
% \begin{tabular}{|l|c|c|c|c|c|c|c|c|}
% \hline
%                   &\begin{tabular}[c]{@{}c@{}}PConv\\ \cite{PConv}\end{tabular}        & \begin{tabular}[c]{@{}c@{}}DF2\\ \cite{yu2019free}\end{tabular}       & \begin{tabular}[c]{@{}c@{}}SC-\\GAN \cite{jo2019sc}\end{tabular}        & \begin{tabular}[c]{@{}c@{}}P2P\\ \cite{isola2017image}\end{tabular}   & Ours  \\ \hline
% Places2       &   0.0\%           & 3.7\%  &18.2\%   &4.5\%   &\textbf{73.6\%}           \\
% CelebA       &     0.2\%      & 0.8\%    & 21.3\%     & 15.1\%    &\textbf{54.2\%}   \\\hline
% \end{tabular}
% \label{tab:user}
% \end{table}

\begin{table*}[t]
\caption{Ablation studies on CelebA-HQ and Places2. We show the numerical results produced by reducing injection times, reducing SGB amounts, removing sketch line guidance, removing texture generation branch, and our complete framework.}
\begin{center}
\begin{tabular}{l|c|c|c|c|cccccc}
\hline
      & \multicolumn{5}{c|}{CelebA-HQ}                                                                        & \multicolumn{5}{c}{Places2}                                                                                                                     \\ \hline
Model & 1 Inject & 1 block & w/o 1-$\sigma$ & \multicolumn{1}{c|}{w/o texture} & \multicolumn{1}{c|}{Ours}           & \multicolumn{1}{c|}{1 Inject} & \multicolumn{1}{c|}{1 block} & \multicolumn{1}{c|}{w/o 1-$\sigma$} & \multicolumn{1}{c|}{w/o texture} & Ours           \\ \hline
PSNR$\uparrow$  & 23.56    & 24.49   & 24.39   & 25.26                            & \multicolumn{1}{c|}{\textbf{25.42}} & \multicolumn{1}{c|}{20.81}    & \multicolumn{1}{c|}{23.55}   & \multicolumn{1}{c|}{23.16}   & \multicolumn{1}{c|}{23.91}       & \textbf{24.30} \\ \hline
SSIM$\uparrow$  & 0.86     & 0.87    & 0.88    & 0.89                             & \multicolumn{1}{c|}{\textbf{0.90}}  & \multicolumn{1}{c|}{0.67}     & \multicolumn{1}{c|}{0.74}    & \multicolumn{1}{c|}{0.72}    & \multicolumn{1}{c|}{0.70}        & \textbf{0.77}  \\ \hline
FID$\downarrow$   & 17.63    & 17.56   & 14.94   & 10.94                            & \multicolumn{1}{c|}{\textbf{9.92}}  & \multicolumn{1}{c|}{105.79}   & \multicolumn{1}{c|}{83.52}   & \multicolumn{1}{c|}{77.80}   & \multicolumn{1}{c|}{69.66}       & \textbf{63.56} \\ \hline

\end{tabular}
\end{center}
\label{tab:ablation}
\vspace{-5mm}
\end{table*}

{\flushleft \bf Numerical Evaluations.}
We evaluate existing methods on the two benchmark datasets numerically from two aspects. First, we use standard metrics including PSNR, SSIM, and FID~\cite{FID} to measure the pixel-level, structure-level, and holistic-level similarities between the output results and original images. When producing these results, we use the sketch lines and color dots from the hole regions without modifications. Table~\ref{tab:all} shows the evaluation results where our method performs favorably against the existing methods. This indicates that our method is more effective to generate user-intended contents while maintaining visual pleasantness. 
%The inference time of PConv, DF2, P2P, FEGAN and ours are 12ms, 27ms, 10ms, 18ms, and 46ms, respectively.

Besides standard metrics, we perform a human subject evaluation. There are over 20 volunteers to evaluate the results on both CelebA-HQ and Places2 datasets. The volunteers all have image processing background. There are 15 rounds for each subject. In each round, the subject needs to select the most visually pleasant result from the 5 results generated by existing methods without knowing the hole region in advance. We tally the votes and show the statistics in the last row in Table~\ref{tab:all}. The comparison results with respect to existing methods indicate that our method is more effective
to generate visually high-quality image content.

\subsection{Ablation Study}
Our DeFLOCNet improves baseline results via SGBs and TGB. We analyze the effects of SGB and TGB on the output results.

\renewcommand{\tabcolsep}{0.5pt}
\def\swfive{0.18\linewidth}
\begin{figure}[t]
    \begin{center}
    \small
\begin{tabular}{ccccc}
\vspace{-0.5mm}\includegraphics[width=\swfive]{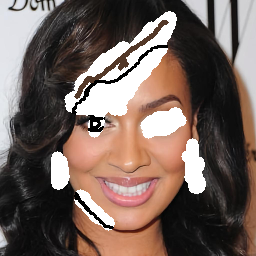}&
\includegraphics[width=\swfive]{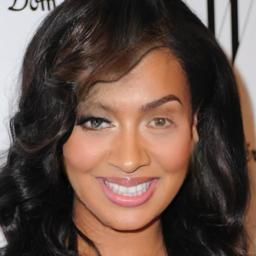}&
\includegraphics[width=\swfive]{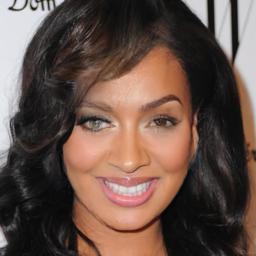}&
\includegraphics[width=\swfive]{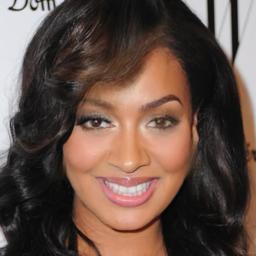}&
\includegraphics[width=\swfive]{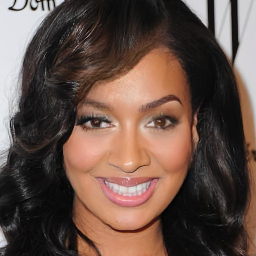}\\

\includegraphics[width=\swfive]{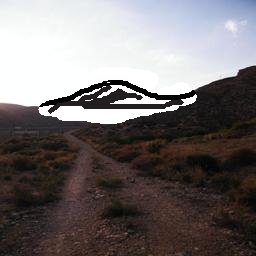}&
\includegraphics[width=\swfive]{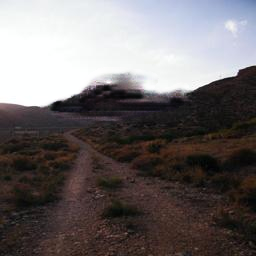}&
\includegraphics[width=\swfive]{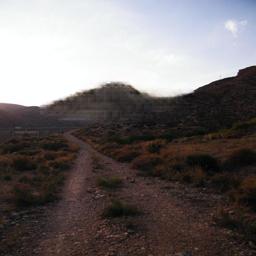}&
\includegraphics[width=\swfive]{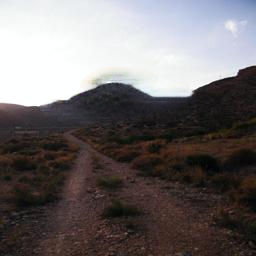}&
\includegraphics[width=\swfive]{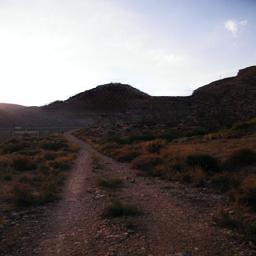}\\
(a) Input & (b) 1 inject & (c) 1 SGB & (d) w/o 1-$\sigma $  & (e)  Ours \\
\end{tabular}
\end{center}
\vspace{-2mm}
\caption{Ablation analysis on SGB. Input images are in (a). The result produced by using only 1 injection in each SGB is in (b), by using only one SGB is in (c), and by not using sketch line guidance (\ie, $1-\sigma$ in Eq.~\ref{eq:color}) is in (d). User intentions are not effectively reflected in these results in comparsion to ours in (e).}
\label{fig:blockabl}
\end{figure}

{\flushleft \bf Structure generation block.}
We set SGBs across multiple skip connection layers to reinforce user intentions from coarse to fine. Meanwhile, within each skip connection layers we gradually enrich encoder feature representations. Fig.~\ref{fig:blockabl} shows two visual examples. Input images are in (a), and we only use 1 injection within each SGB to produce the results in (b). On the other side, we only use one SGB set on the middle-level skip connection layer, and generate the results in (c). The results in (b) and (c) indicate that using limited injections or SGBs are not effective during structure generation. In constrast to these two configurations, we do not use sketch line constraint and produce the results in (d). They show that color propagates in arbitrary directions for unintended structure generation. By using more injections and propagation guidance, we are able to produce visually pleasant structures in (e). Table~\ref{tab:ablation} numerically shows that multiple injections and propagation constraints improve the structure quality of the generated content.

\def\swthree{0.2\linewidth}
\renewcommand{\tabcolsep}{0.5pt}
\begin{figure}
    \begin{center}
    \small
\begin{tabular}{ccc}
\vspace{-0.5mm}\includegraphics[width=\swthree]{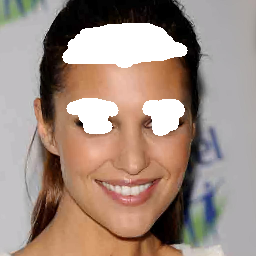}&
\includegraphics[width=\swthree]{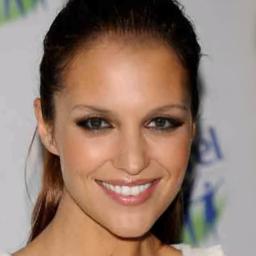}&
\includegraphics[width=\swthree]{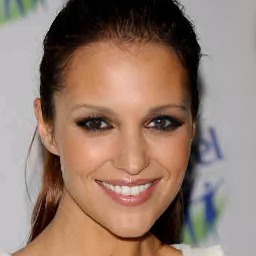}\\
\includegraphics[width=\swthree]{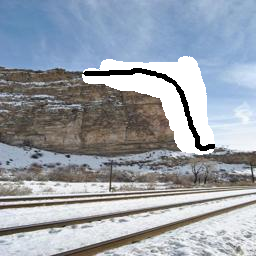}&
\includegraphics[width=\swthree]{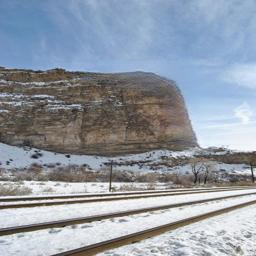}&
\includegraphics[width=\swthree]{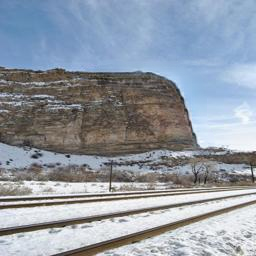}\\
(a) Input &(b) w/o TGB  &(c) w/ TGB\\
\end{tabular}
\end{center}
\vspace{-2mm}
\caption{Ablation analysis on TGB. Inputs are in (a). The results of our method without using TGB are shown in (b). The results produced by using TGB are shown in (c) where there are more textures and details.}
\label{fig:textureabl}
\end{figure}

{\flushleft \bf Texture generation branch.} 
We analyze the effect of TGB by comparing the results produced with TGB and without TGB. Fig.~\ref{fig:textureabl} shows two examples. Input images with user controls in (a). Without TGB, texture details are blurry in some regions in (b) (\eg, the forelock hair and mountain boundaries). The utilization of TGB synthesizes texture details based on the input image and thus reduces the blurry artifacts in (c). The numerical evaluation in Table~\ref{tab:ablation} also indicates that TGB improves the generated content.

% There is a long existing dilemma in image editing to achieve both control flexibility and user intention accuracy. Existing attempts tend to attach low-level controls into color images for a deep encoder-decoder input. This configuration diminishes user intentions gradually as the encoder features do not effectively represent such diverged data.
\section{Concluding Remarks}
 We propose structure generation blocks set on skip connection layers to receive low-level controls, while only color images are sent to the encoder. The encoder features, representing only color images, are modulated via these blocks gradually to reinforce user intentions within each skip connection layer. Furthermore, the modulated encoder features with structure ingredients supplement the decoder features together with the generated texture features across multiple skip connection layers. Therefore, both structures and textures are generated from coarse to fine in the CNN feature space, bringing both user-intended and visually pleasing image content. The Experiments on the benchmark datasets indicate the effectiveness of our methods both numerically and visually compared to the state-of-the-art approaches.
\section{Acknowledgement}
This work was supported in part by the National Natural Science Foundation of China under grant 62072169.

This work was partially supported by an ECS grant from the Research Grants Council of the Hong Kong (Project No. CityU 21209119) and an APRC grant from CityU, Hong Kong (Project No. 9610488) .
\newpage
{\small

\bibliographystyle{ieee_fullname}
% \bibliography{egbib}
}

\end{document}